\documentclass[sigconf]{acmart}

\AtBeginDocument{%
  \providecommand\BibTeX{{%
    \normalfont B\kern-0.5em{\scshape i\kern-0.25em b}\kern-0.8em\TeX}}}

\acmConference[AdvML '20]{AdvML '20: Workshop on Adversarial Learning Methods for Machine Learning and Data Mining}{August 24, 2020}{San Diego, CA}
\acmBooktitle{AdvML '20: Workshop on Adversarial Learning Methods for Machine Learning and Data Mining, August 24, 2020, San Diego, CA}

\usepackage{algorithm}
\usepackage{algpseudocode}
\begin{document}

\title{Garment Design with Generative Adversarial Networks}


\author{Chenxi Yuan}
\affiliation{\institution{Northeastern University}}
\email{yuan.chenx@northeastern.edu}

\author{Mohsen Moghaddam}
\affiliation{\institution{ Northeastern University}}
\email{mohsen@northeastern.edu}

\renewcommand{\shortauthors}{Yuan and Moghaddam}

\begin{abstract}
The designers' tendency to adhere to a specific mental set and heavy emotional investment in their initial ideas often hinder their ability to innovate during the design thinking and ideation process. In the fashion industry, in particular, the growing diversity of customers' needs, the intense global competition, and the shrinking time-to-market (a.k.a., ``fast fashion'') further exacerbate this challenge for designers. Recent advances in deep generative models have created new possibilities to overcome the cognitive obstacles of designers through automated generation and/or editing of design concepts. This paper explores the capabilities of generative adversarial networks (GAN) for automated attribute-level editing of design concepts. Specifically, attribute GAN (AttGAN)---a generative model proven successful for attribute editing of human faces---is utilized for automated editing of the visual attributes of garments and tested on a large fashion dataset. The experiments support the hypothesized potentials of GAN for attribute-level editing of design concepts, and underscore several key limitations and research questions to be addressed in future work.
\end{abstract}




\maketitle

\section{Introduction}
Technology-driven innovation through AI and machine learning has become an essential success factor for fashion design firms in the 21st century. According to McKinsey \& Company \cite{Amed2018}, over 140\% of the global fashion industry profit is generated by the leading 20\% of the fashion brands. As a result, significant recent progress has been made in adopting AI and machine learning techniques for augmented and personalized design. Examples range from style matching \cite{kalantidis2013getting,liu2016deepfashion,xu2019evading}, to trend forecasting \cite{al2017fashion}, interactive search \cite{zhao2017memory,kovashka2012whittlesearch}, style recommendation \cite{lin2018explainable,simo2015neuroaesthetics}, virtually trying clothes on \cite{han2018viton}, and clothing type and style classification \cite{liang2016clothes,zhu2017your}. AI and machine learning research in the fashion industry has the promise to directly influence the purchasing behavior of customers, the garment design thinking and ideation process, user-centered design and mass-personalization, and the ability of the fashion industry to adapt their product development strategies accordingly.

This article investigates how generative adversarial networks (GAN) \cite{goodfellow2014generative} can be adopted to enabled automated attribute-level editing of past successful products to inform new product design and development processes. Different from conventional adversarial attacks \cite{xu2018structured,xu2019interpreting,xu2019topology}, attribute editing involves making translations/adjustments to images based on the target attributes to generate a new sample with desired attributes while preserving other original details. Current GAN-based attribute editing research is predominantly centered on human face images \cite{ehrlich2016facial,schroff2015facenet}. The facial attribute editing task allows to edit a face image by manipulating single or multiple attributes of interest such as hair color, expression, mustache, and age \cite{he2019attgan}. For fashion products, the analogous visual attributes of interest may include style type, sleeve length, color, and pattern, among others. The ability to manipulate the attributes of a prior design is particularly useful in a variety of situations where customers are not satisfied with certain attributes or would like to explore various combinations of them \cite{zhu2017your}.

Conditional GAN \cite{mirza2014conditional} is an extension of the original GAN formulation which allows to generate images conditioned on user-defined features that control the generative process. Among the various versions of conditional GAN proposed to date \cite{reed2016generative,perarnau2016invertible,kaneko2017generative}, attribute GAN (AttGAN) \cite{he2019attgan} has proven effective in generating realistic edited images with desired attributes on human face dataset. AttGAN can generate visually more pleasing results with fine facial details in comparison with the state-of-the-art GAN models. However, there is no proof or indication that AttGAN can be directly applied for attribute-level editing of fashion data such as garment images with acceptable performance. 

To tackle this problem, this article develops and tests a novel AttGAN model that enables attribute-level editing of fashion items while preserving other visual aspects and attributes. First, the original AttGAN model \cite{he2019attgan} is implemented on a large fashion dataset consisting of 13,221 garment images along with 22 attribute values. Numerical experiments are then conducted to edit the images with respect to five desired attributes including ``vest'', ``polo'', ``hoodie'', ``blouse'' and ``T-shirt'' (e.g., selecting the attribute ``vest'' is desired to turn any type of shirt into a vest). The experiments show that the great performance of AttGAN on the human face editing task cannot be achieved on the fashion editing task. The authors hypothesize the underlying reason to stem from the \emph{relative area} of editing which, unlike human faces, corresponds to a large area of a garment image (e.g., entire sleeve or collar). A new version of AttGAN is thus developed to address this limitation. Numerical experiments indicate significant improvement in successful editing of different attributes such as sleeve length, color, pattern and clothes type, while preserving the remainder of the original garment image.

\section{Related Work}
Since its introduction in 2014, GAN \cite{goodfellow2014generative} continues to attract growing interests in the deep learning community and has been applied to various domains such as computer vision, natural language processing \cite{nie2018relgan,che2017maximum}, time series synthesis \cite{esteban2017real,luo2018multivariate}, and semantic segmentation \cite{luc2016semantic}. Specifically, GAN has shown significant recent success in the field of computer vision field on a variety of tasks such as image generation \cite{choe2017face,zhang2017stackgan}, image to image translation \cite{zhu2017toward,isola2017image}, and image super-resolution \cite{ledig2017photo,dong2017semantic}, among others. The standard GAN structure comprises two neural networks: a generator ${G}$ and a discriminator ${D}$ iteratively trained by competing against each other in a minimax game with the following learning objective:

\begin{equation}
\min _{G} \max _{D} \mathbb{E}_{\mathbf{x} \sim p_{\text {data}}}[\log D(\mathbf{x})]+\mathbb{E}_{\mathbf{z} \sim p_{\mathbf{z}}}[\log (1-D(G(\mathbf{z})))],
\end{equation}
where ${\mathbf{z}}$ is a random or encoded vector, $p_{\text {data}}$ is the empirical distribution of the input training images, and $p_{\mathbf{z}}$ is the prior distribution of ${\mathbf{z}}$ (e.g., normal distribution).

In the standard GAN model, there is no control over the modes of the data being generated. In conditional GAN (cGAN) \cite{mirza2014conditional}, however, the generative process is conditioned to generate images based on a user-defined vector of features. The learning objective of cGAN is as follows:

\begin{equation}
\begin{split}
    \min _{G} \max _{D} \mathbb{E}_{\mathbf{x,y} \sim p_{\text {data}}}[\log D(\mathbf{x,b})]+\\ \mathbb{E}_{\mathbf{z} \sim p_{\mathbf{z, b}} \sim p_{\mathbf{b}}}\left[\log \left(1- D\left(G\left(\mathbf{z,b}\right), \mathbf{b}\right)\right)\right],
\end{split}
\end{equation} 
where ${\mathbf{b}}$ is the extra information (e.g., class labels, attribute information) for a given real sample ${\mathbf{x}}$ as input. cGAN allows to control the generation of samples using $\mathbf{b}$. 

In cGAN, the generation of samples can be conditioned on class information \cite{odena2017conditional} , text description \cite{reed2016generative,zhang2018generative,xu2018attngan}, audio \cite{chen2019hierarchical,chen2017deep}, skeleton \cite{ma2017pose,raj2018swapnet}, and attributes \cite{shen2017learning}. In the fashion industry, researchers have applied GANs for a variety of applications such as: (i) automated garment textures filling \cite{xian2018texturegan}, (ii) texture transferring \cite{jiang2017fashion} where given a basic clothing image and a fashion style image, (iii) virtual try-on \cite{zhu2017your} aimed at creating new clothing on a human body based on textual descriptions, (iv) interactive image editing \cite{cheng2018sequential} where users can guide an agent to edit images via multi-turn conversational language, and (v) fashion recommendation \cite{kang2017visually} in which the model can be used for both personalized recommendation and personalized fashion design.

\section{Methodology}
This section first introduces the original formulation of a cGAN for attribute-level editing: attribute GAN (AttGAN). AttGAN \cite{he2019attgan} has shown great performance on facial image editing with binary attributes (e.g., \{mustache, no-mustache\}) and is used as our baseline model. Next, an in-depth analysis of the AttGAN formulation is conducted and a modified version of AttGAN is developed to achieve comparable performance on garment image editing.

\subsection{Attribute Generative Adversarial Networks}
\label{S3-1}
A limitation of conventional cGAN is that the user-defined attributes/labels affect the editing of the entire image including the parts unrelated to the desired attribute. To avoid this limitation, AttGAN \cite{he2019attgan} builds an effective framework for high quality facial attribute editing and simultaneously preserving attribute-excluding details.

The learning objectives of the AttGAN generator and AttGAN discriminator and classifier are respectively as follows: 
\begin{equation}\label{eq:overall_G}
\min _{G_{enc}, G_{dec}} \mathcal{L}_{enc, dec}=\lambda_{1} \mathcal{L}_{rec}+\lambda_{2} \mathcal{L}_{cls_{g}}+\mathcal{L}_{adv_{g}},
\end{equation}
\begin{equation}\label{eq:overall_D_C}
\min _{D, C} \mathcal{L}_{dis, cls}=\lambda_{3} \mathcal{L}_{c l s_{c}}+\mathcal{L}_{adv_{d}},
\end{equation}
where $\mathcal{L}_{rec}$ is the reconstruction loss for satisfactory preservation of attribute-excluding details, $\mathcal{L}_{cls}$ is the classification constraint to guarantee the correct editing of the desired attributes, and $\mathcal{L}_{adv}$ is the adversarial learning employed for visually realistic editing. $\lambda_{1}, \lambda_{2}$ and $\lambda_{3}$ are hyperparameters that control the importance of different terms and are tuned experimentally.

Inspired by AttGAN's success in human facial attribute editing, the authors first attempted to utilize the original AttGAN model for attribute-level editing of fashion product images. The preliminary observations was that AttGAN model does not perform as expected on fashion data such as garment images. Specifically, the observation was that although AttGAN can reconstruct original fashion images, it is unable to generate new vivid image with the desired attributes modified. The underlying reason behind such poor performance on fashion data is elaborated and address in the remainder of this section. 

In Eq. ~\eqref{eq:overall_G},  $\mathcal{L}_{cls_{g}}$ is the attribute classification loss, employed to guide the generative process to learn and edit the desired attributes. The reconstruction loss $\mathcal{L}_{rec}$, on the other hand, is intended to enable the decoder to reconstruct the original input images so that the generated samples can preserve the attributes-excluding details. In the original AttGAN, these two loss functions are both trained on the generator. The aforementioned problem contributing to the poor performance of AttGAN on fashion data stems from an inherent conflict between these two loss functions. The classification loss wants the generator to distinguish the desired attributes $\mathbf{b}$ from the original images $\mathbf{x}^{\mathbf{a}}$, by minimizing the summation of the binary cross entropy of the desired attributes and input images as follows:
\begin{equation}\label{eq:clas_b}
\min _{G_{enc}, G_{dec}} \mathcal{L}_{cls_{g}}=\mathbb{E}_{\mathbf{x}^{\mathbf{a}} \sim p_{data}, \mathbf{b} \sim p_{attr}}\left[\ell_{g}\left(\mathbf{x}^{\mathbf{a}}, \mathbf{b}\right)\right],
\end{equation} 
\begin{equation}
\mbox{\small $\ell_{g}\left(\mathbf{x}^{\mathbf{a}}, \mathbf{b}\right)={ \sum_{i=1}^{n}} -b_{i} \log C_{i}(\mathbf{x}^{\hat{\mathbf{b}}})-(1-b_{i}) \log (1-C_{i}( \mathbf{x}^{\hat{\mathbf{b}}}))$},
\end{equation} 
where $\mathbf{x}^{\hat{\mathbf{b}}}$ is the edited image expected to change the attributes of $\mathbf{x}^{\mathbf{a}}$ to another attributes $\mathbf{b}$. This is achieved by decoding latent representation $\mathbf{z}$ conditioned on attributes $\mathbf{b}$: $\mathbf{x}^{\hat{\mathbf{b}}} = G_{dec}(\mathbf{z}, \mathbf{b})$, where $\mathbf{z}$ is encoded from image $\mathbf{x}^{\mathbf{a}}$ with $n$ binary attributes $\mathbf{a}$ and is denoted by $\mathbf{z}=G_{enc}\left(\mathbf{x}^{\mathbf{a}}\right)$. Therefore the generated image is formulated as $\mathbf{x}^{\hat{\mathbf{b}}} =G_{dec}(G_{enc}\left(\mathbf{x}^{\mathbf{a}}\right), \mathbf{b})$.

The reconstruction loss, on the other hand, wants the generator to preserve the original images as much as possible, by minimizing the following Manhattan distance function:
\begin{equation}\label{eq:rec}
\min _{G_{enc}, G_{dec}} \mathcal{L}_{rec}=\mathbb{E}_{\mathbf{x}^{a} \sim p_{data}}\left[\left\|\mathbf{x}^{\mathbf{a}}-\mathbf{x}^{\hat{\mathbf{a}}}\right\|_{1}\right]
\end{equation}
where $\mathbf{x}^{\hat{\mathbf{a}}} =G_{dec}(G_{enc}\left(\mathbf{x}^{\mathbf{a}}\right), \mathbf{a})$. The reconstruction loss enables the decoder to restore the original images conditioned on its own attributes $\mathbf{a}$ from $\mathbf{z}$.

The aforementioned conflict limits the use of the AttGAN for attribute-level editing of fashion data where some attributes account for larger relative areas of the image (e.g., entire sleeve or collar). It was observed that the classification loss is unable generate distinct samples from the original images of garment. The AttGAN model is thus reformulated by taking the classification loss out and training it on the generator independently. This, in turn, would enable more flexibility for the generator to edit larger areas of the input image. Accordingly, Eq.~\eqref{eq:overall_G} is recast as follows 

\begin{equation}\label{eq:G_1}
\min _{G_{enc}, G_{dec}} \mathcal{L}_{enc, dec}=\lambda_{1}\mathcal{L}_{rec}+\mathcal{L}_{adv_{g}},
\end{equation}
\begin{equation}\label{eq:G_2}
\min _{G_{enc}, G_{dec}} \mathcal{L}_{enc, dec}= \lambda_{2} \mathcal{L}_{cls_{g}}.
\end{equation}
The modified model is referred to as Design-AttGAN with the training procedure elaborated in ~\ref{alg: D-AttGAN}.

\begin{algorithm}
\caption{Design-AttGAN}
\begin{algorithmic}[1]
\State \textbf{Input:} images $\mathbf{X}$ and their attributes $\mathbf{A}$ and step number $N$
\For{step $\gets$ 0 to $N$}
        \State Sample batch $\mathbf{x}^{\mathbf{a}} \in \mathbf{X}$,  $\mathbf{a} \in \mathbf{A}$ and random generate $\mathbf{b}$ 
        \For{inner step $\gets$ 0 to 5}
                \State Minimize Eq.\eqref{eq:overall_D_C}
        \EndFor
        \State Minimize Eq.\eqref{eq:G_1}
        \State Minimize Eq.\eqref{eq:G_2}
\EndFor
\State \textbf{Output:} a well-trained $G_{enc}$ and $G_{dec}$
\end{algorithmic}\label{alg: D-AttGAN}
\end{algorithm}

\section{Experiments}
\label{S4}

\subsection{Dataset \& Training}
\label{S4-1}
The Design-AttGAN model is tested on a fashion dataset \cite{ping2019fashion}, which contains 13221 images and each of which has annotation of 22 binary attributes (with/without). Attributes with great frequency are chosen in all our experiments, including ``vest'', ``polo'', ``stripe'', ``short sleeve'', ``long sleeve'', ``red'', ``yellow'', ``blue'', ``purple'', ``black'', and ``white''. The dataset is separated into training set for training model and testing set for evaluation. The experiments are conducted on TensorFlow using the open-source code provided by \cite{he2019attgan}. The model is trained by Adam optimizer with the batch size of 32 and the learning rate of 0.0002. 

\begin{figure}
    \centering
    \includegraphics[width=0.3\textwidth, height=50mm]{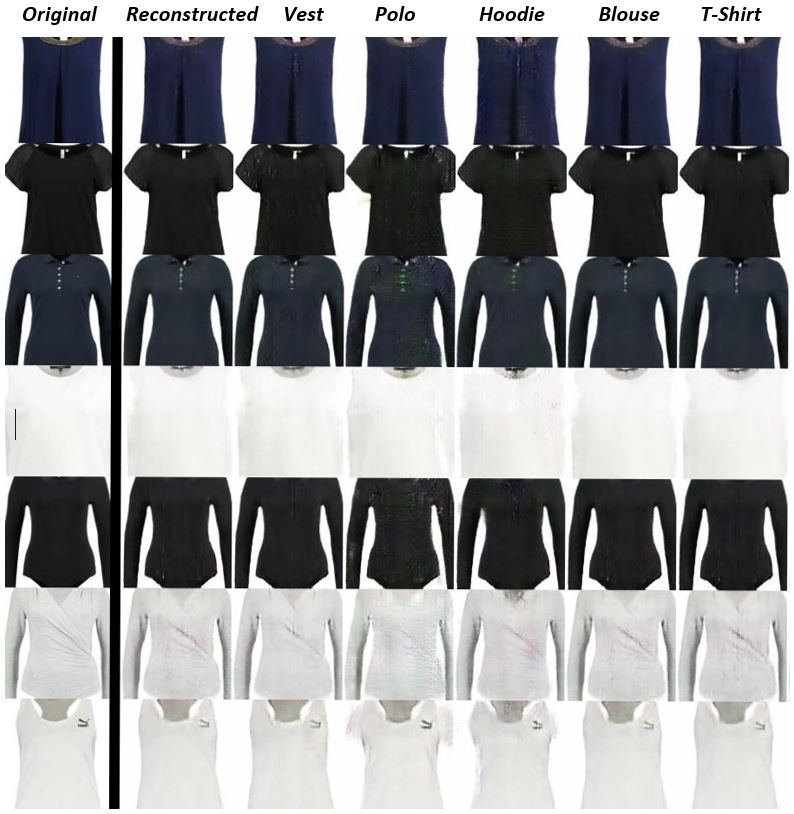}
    \caption{AttGAN on fashion images.}
    \label{fig:att}
\end{figure}

\begin{figure}
    \centering
    \includegraphics[width=0.3\textwidth,height=50mm]{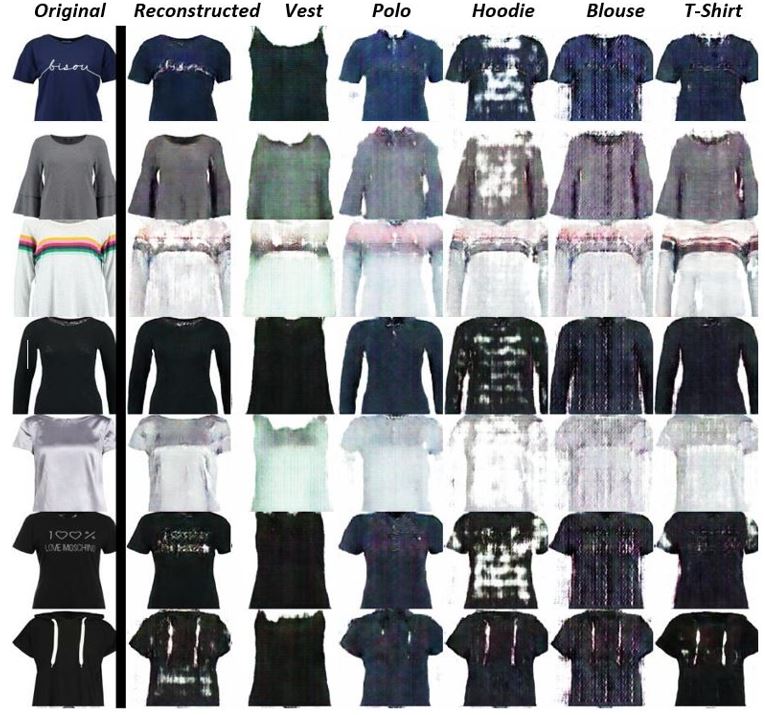}
    \caption{Design-AttGAN on fashion images.}
    \label{fig:f_att}
\end{figure}

 \begin{figure*}
    \centering
    \includegraphics[width=0.85\textwidth]{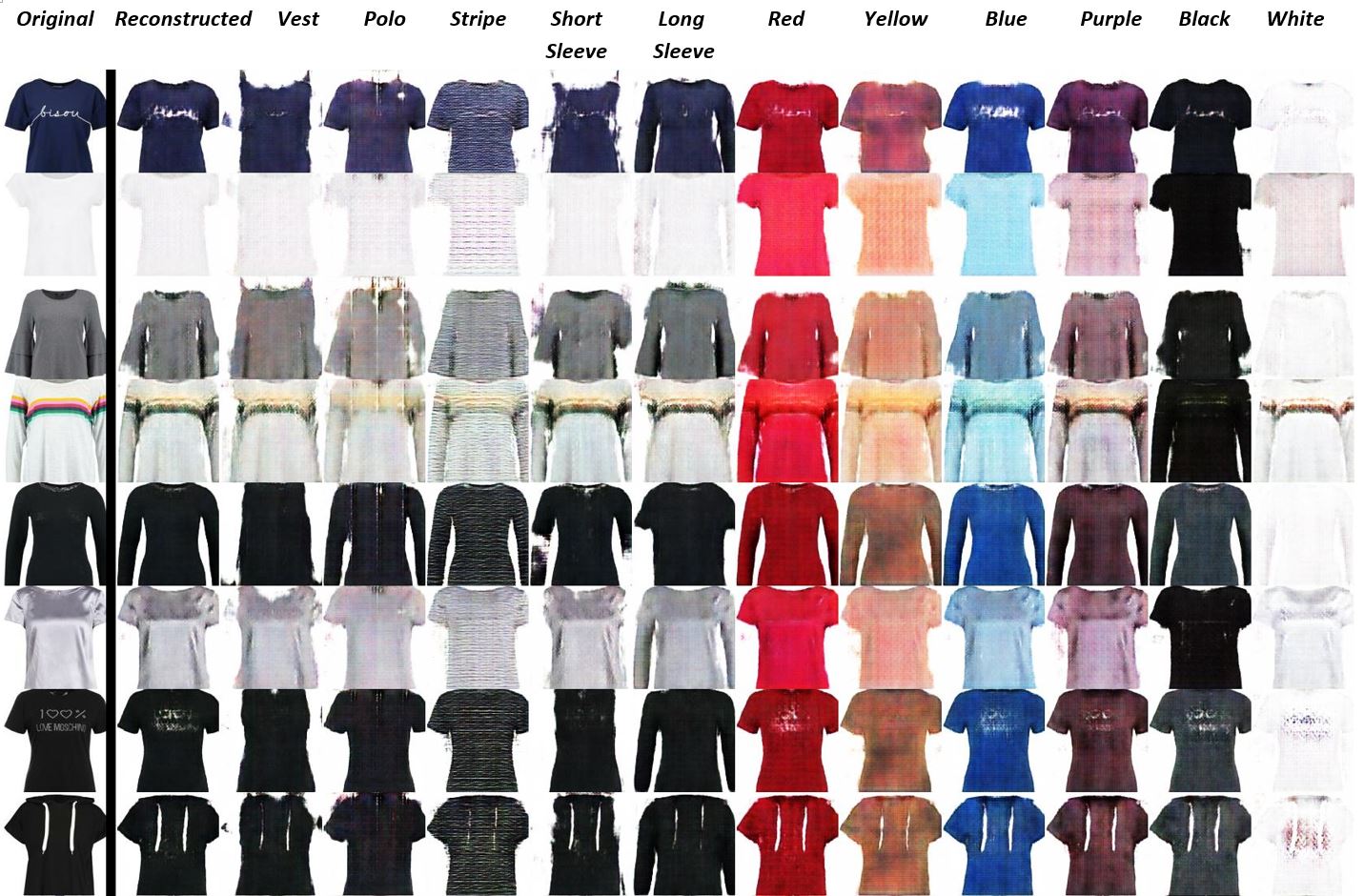}
    \caption{Design-AttGAN for different attribute editing tasks.}
    \label{fig:f_att_more}
\end{figure*}

\subsection{Results}
\label{S4-2}
Figure \ref{fig:att} shows the binary attribute editing results obtained from the original AttGAN model. As can be seen, AttGAN performs poorly on the garment images. The model dose not preserve any garment patterns and is not even able to properly edit. The underlying reason is that the classification learning task is negatively influenced by the reconstruction learning task in the original AttGAN formulation. The Design-AttGAN model is applied to address this problem (Figure \ref{fig:f_att}). In the Design-AttGAN model, the classification loss is trained as an independent objective function to enhance the ability of the generator for attribute editing. As Figure \ref{fig:f_att} shows, the Design-AttGAN model outperforms the baseline AttGAN model in learning multiple attributes and changing the type of garment to ``vest'' or ``polo''.

In the Design-AttGAN model, the classification loss $\mathcal{L}_{cls_{g}}$ is trained separately without the restriction of the reconstruction and adversarial losses in the minimax game. This provides the model with more flexibility to generate more good ``fake'' samples. This is necessary for the fashion attribute editing task because unlike the facial attributes, the attributes of garment products typically account for a relatively larger area of the image. This way, the generative model would have to generate more ``wild'' samples to incorporate the edited attributes in the original images.

\paragraph{Attributes Extension} 
Figure \ref{fig:f_att_more} shows the performance of the Design-AttGAN model on the attribute-editing task with eleven distinct attributes, including clothe type (``vest'', ``polo''), clothe pattern (``slim horizontal stripes''), length of sleeves (``short sleeves'', ``long sleeves''), and multiple colors (``red'', ``yellow'', ``blue'', ``purple'', ``black'', ``white''). As shown, the model can successfully edit the color and stripe on clothes, and change the length of sleeves. However, it is not able to learn the latent pattern associated with the ``polo'' attribute. 

\begin{figure}
    \centering
    \begin{tabular}{c|c}
    \includegraphics[width=0.18\textwidth,height=60mm]{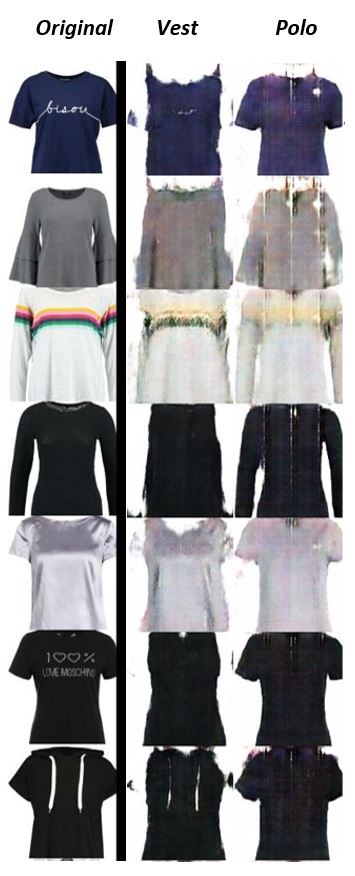}
    &\includegraphics[width=0.18\textwidth,height=60mm]{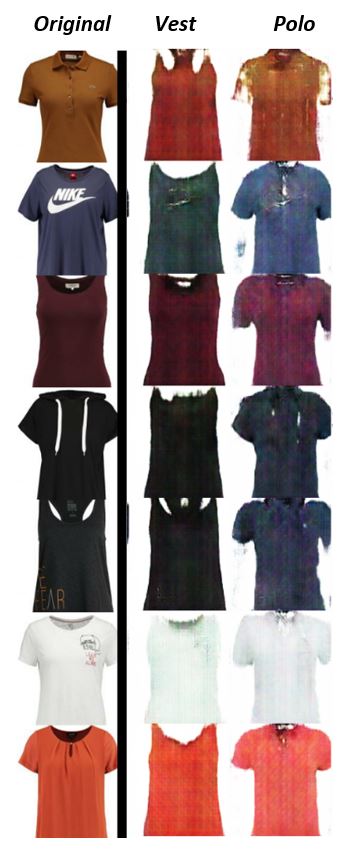}
    \end{tabular}
    \caption{Design-AttGAN trained on the entire dataset (left) and narrowed dataset (right). }
    \label{fig:my_label}
\end{figure}

To solve this problem, another experiment was conducted on a \emph{narrowed} dataset. The original dataset has over 12,000 images; however, unlike sleeve length and color that are the indispensable  attributes of any garment type, attributes such as ``polo'' are relatively rare and thus cause the data to be imbalanced. Hence, 5,782 images with attributes of clothe type category (e.g., vest, polo, blouse, t-shirt, hoodie, one-piece dress) are picked up from the original dataset. The Design-AttGAN model is then retrained on this narrowed dataset to generate samples with the desired attributes ``vest'' and ``polo''. Results showed that the Design-AttGAN model yields better performance on the narrowed dataset (Figure \ref{fig:my_label}, right) than the original dataset (Figure \ref{fig:my_label}, left). With the narrowed dataset where each image is guaranteed to contain clothes type attributes, the model is more likely to capture these attributes and edit accordingly. This is further proof to the fact that training a cGAN model is highly sensitive to the balance of different attributes in the dataset. The imbalanced distribution of attributes hinders the ability of the model to learn the attributes with low frequency in the dataset.  

\section{Conclusions and Future Work}

This paper introduces a deep learning model, Design-AttGAN, which has the ability to automatically edit garment images conditioned on certain user-defined attributes. The performance of the generative model is experimented and tested on a large fashion dataset. The original formulation of the AttGAN is modified to avoid the inherent conflict between the reconstruction loss and the attribute classification loss. An important observation was that generative adversarial networks are sensitive to different domains and therefore need careful revision and hand-engineering of the algorithms based on the dataset and target task.

Future work will involve seeking to generate images with higher resolution, improve the stability of the Design-AttGAN model, and broaden the scope of the proposed methodology. Evaluation of GAN's performance will also be an important area to explore.

\newpage
\bibliographystyle{ACM-Reference-Format}
\bibliography{sample-base}


\begin{thebibliography}{47}


\ifx \showCODEN    \undefined \def \showCODEN     #1{\unskip}     \fi
\ifx \showDOI      \undefined \def \showDOI       #1{#1}\fi
\ifx \showISBNx    \undefined \def \showISBNx     #1{\unskip}     \fi
\ifx \showISBNxiii \undefined \def \showISBNxiii  #1{\unskip}     \fi
\ifx \showISSN     \undefined \def \showISSN      #1{\unskip}     \fi
\ifx \showLCCN     \undefined \def \showLCCN      #1{\unskip}     \fi
\ifx \shownote     \undefined \def \shownote      #1{#1}          \fi
\ifx \showarticletitle \undefined \def \showarticletitle #1{#1}   \fi
\ifx \showURL      \undefined \def \showURL       {\relax}        \fi
\providecommand\bibfield[2]{#2}
\providecommand\bibinfo[2]{#2}
\providecommand\natexlab[1]{#1}
\providecommand\showeprint[2][]{arXiv:#2}

\bibitem[\protect\citeauthoryear{Al-Halah, Stiefelhagen, and Grauman}{Al-Halah
  et~al\mbox{.}}{2017}]%
        {al2017fashion}
\bibfield{author}{\bibinfo{person}{Ziad Al-Halah}, \bibinfo{person}{Rainer
  Stiefelhagen}, {and} \bibinfo{person}{Kristen Grauman}.}
  \bibinfo{year}{2017}\natexlab{}.
\newblock \showarticletitle{Fashion forward: Forecasting visual style in
  fashion}. In \bibinfo{booktitle}{\emph{Proceedings of the IEEE International
  Conference on Computer Vision}}. \bibinfo{pages}{388--397}.
\newblock


\bibitem[\protect\citeauthoryear{Amed, Andersson, Berg, Drageset, Hedrich, and
  Kappelmark}{Amed et~al\mbox{.}}{2018}]%
        {Amed2018}
\bibfield{author}{\bibinfo{person}{Imran Amed}, \bibinfo{person}{Johanna
  Andersson}, \bibinfo{person}{Achim Berg}, \bibinfo{person}{Martine Drageset},
  \bibinfo{person}{Saskia Hedrich}, {and} \bibinfo{person}{Sara Kappelmark}.}
  \bibinfo{year}{2018}\natexlab{}.
\newblock \bibinfo{title}{{The State of Fashion: Renewed optimism for the
  fashion industry}}.
\newblock
\newblock
\urldef\tempurl%
\url{https://www.mckinsey.com/industries/retail/our-insights/renewed-optimism-for-the-fashion-industry{\#}}
\showURL{%
\tempurl}


\bibitem[\protect\citeauthoryear{Che, Li, Zhang, Hjelm, Li, Song, and
  Bengio}{Che et~al\mbox{.}}{2017}]%
        {che2017maximum}
\bibfield{author}{\bibinfo{person}{Tong Che}, \bibinfo{person}{Yanran Li},
  \bibinfo{person}{Ruixiang Zhang}, \bibinfo{person}{R~Devon Hjelm},
  \bibinfo{person}{Wenjie Li}, \bibinfo{person}{Yangqiu Song}, {and}
  \bibinfo{person}{Yoshua Bengio}.} \bibinfo{year}{2017}\natexlab{}.
\newblock \showarticletitle{Maximum-likelihood augmented discrete generative
  adversarial networks}.
\newblock \bibinfo{journal}{\emph{arXiv preprint arXiv:1702.07983}}
  (\bibinfo{year}{2017}).
\newblock


\bibitem[\protect\citeauthoryear{Chen, Maddox, Duan, and Xu}{Chen
  et~al\mbox{.}}{2019}]%
        {chen2019hierarchical}
\bibfield{author}{\bibinfo{person}{Lele Chen}, \bibinfo{person}{Ross~K Maddox},
  \bibinfo{person}{Zhiyao Duan}, {and} \bibinfo{person}{Chenliang Xu}.}
  \bibinfo{year}{2019}\natexlab{}.
\newblock \showarticletitle{Hierarchical cross-modal talking face generation
  with dynamic pixel-wise loss}. In \bibinfo{booktitle}{\emph{Proceedings of
  the IEEE Conference on Computer Vision and Pattern Recognition}}.
  \bibinfo{pages}{7832--7841}.
\newblock


\bibitem[\protect\citeauthoryear{Chen, Srivastava, Duan, and Xu}{Chen
  et~al\mbox{.}}{2017}]%
        {chen2017deep}
\bibfield{author}{\bibinfo{person}{Lele Chen}, \bibinfo{person}{Sudhanshu
  Srivastava}, \bibinfo{person}{Zhiyao Duan}, {and} \bibinfo{person}{Chenliang
  Xu}.} \bibinfo{year}{2017}\natexlab{}.
\newblock \showarticletitle{Deep cross-modal audio-visual generation}. In
  \bibinfo{booktitle}{\emph{Proceedings of the on Thematic Workshops of ACM
  Multimedia 2017}}. \bibinfo{pages}{349--357}.
\newblock


\bibitem[\protect\citeauthoryear{Cheng, Gan, Li, Liu, and Gao}{Cheng
  et~al\mbox{.}}{2018}]%
        {cheng2018sequential}
\bibfield{author}{\bibinfo{person}{Yu Cheng}, \bibinfo{person}{Zhe Gan},
  \bibinfo{person}{Yitong Li}, \bibinfo{person}{Jingjing Liu}, {and}
  \bibinfo{person}{Jianfeng Gao}.} \bibinfo{year}{2018}\natexlab{}.
\newblock \showarticletitle{Sequential attention gan for interactive image
  editing via dialogue}.
\newblock \bibinfo{journal}{\emph{arXiv preprint arXiv:1812.08352}}
  (\bibinfo{year}{2018}).
\newblock


\bibitem[\protect\citeauthoryear{Choe, Park, Kim, Hyun~Park, Kim, and
  Shim}{Choe et~al\mbox{.}}{2017}]%
        {choe2017face}
\bibfield{author}{\bibinfo{person}{Junsuk Choe}, \bibinfo{person}{Song Park},
  \bibinfo{person}{Kyungmin Kim}, \bibinfo{person}{Joo Hyun~Park},
  \bibinfo{person}{Dongseob Kim}, {and} \bibinfo{person}{Hyunjung Shim}.}
  \bibinfo{year}{2017}\natexlab{}.
\newblock \showarticletitle{Face generation for low-shot learning using
  generative adversarial networks}. In \bibinfo{booktitle}{\emph{Proceedings of
  the IEEE International Conference on Computer Vision Workshops}}.
  \bibinfo{pages}{1940--1948}.
\newblock


\bibitem[\protect\citeauthoryear{Dong, Yu, Wu, and Guo}{Dong
  et~al\mbox{.}}{2017}]%
        {dong2017semantic}
\bibfield{author}{\bibinfo{person}{Hao Dong}, \bibinfo{person}{Simiao Yu},
  \bibinfo{person}{Chao Wu}, {and} \bibinfo{person}{Yike Guo}.}
  \bibinfo{year}{2017}\natexlab{}.
\newblock \showarticletitle{Semantic image synthesis via adversarial learning}.
  In \bibinfo{booktitle}{\emph{Proceedings of the IEEE International Conference
  on Computer Vision}}. \bibinfo{pages}{5706--5714}.
\newblock


\bibitem[\protect\citeauthoryear{Ehrlich, Shields, Almaev, and Amer}{Ehrlich
  et~al\mbox{.}}{2016}]%
        {ehrlich2016facial}
\bibfield{author}{\bibinfo{person}{Max Ehrlich}, \bibinfo{person}{Timothy~J
  Shields}, \bibinfo{person}{Timur Almaev}, {and} \bibinfo{person}{Mohamed~R
  Amer}.} \bibinfo{year}{2016}\natexlab{}.
\newblock \showarticletitle{Facial attributes classification using multi-task
  representation learning}. In \bibinfo{booktitle}{\emph{Proceedings of the
  IEEE Conference on Computer Vision and Pattern Recognition Workshops}}.
  \bibinfo{pages}{47--55}.
\newblock


\bibitem[\protect\citeauthoryear{Esteban, Hyland, and R{\"a}tsch}{Esteban
  et~al\mbox{.}}{2017}]%
        {esteban2017real}
\bibfield{author}{\bibinfo{person}{Crist{\'o}bal Esteban},
  \bibinfo{person}{Stephanie~L Hyland}, {and} \bibinfo{person}{Gunnar
  R{\"a}tsch}.} \bibinfo{year}{2017}\natexlab{}.
\newblock \showarticletitle{Real-valued (medical) time series generation with
  recurrent conditional gans}.
\newblock \bibinfo{journal}{\emph{arXiv preprint arXiv:1706.02633}}
  (\bibinfo{year}{2017}).
\newblock


\bibitem[\protect\citeauthoryear{Goodfellow, Pouget-Abadie, Mirza, Xu,
  Warde-Farley, Ozair, Courville, and Bengio}{Goodfellow et~al\mbox{.}}{2014}]%
        {goodfellow2014generative}
\bibfield{author}{\bibinfo{person}{Ian Goodfellow}, \bibinfo{person}{Jean
  Pouget-Abadie}, \bibinfo{person}{Mehdi Mirza}, \bibinfo{person}{Bing Xu},
  \bibinfo{person}{David Warde-Farley}, \bibinfo{person}{Sherjil Ozair},
  \bibinfo{person}{Aaron Courville}, {and} \bibinfo{person}{Yoshua Bengio}.}
  \bibinfo{year}{2014}\natexlab{}.
\newblock \showarticletitle{Generative adversarial nets}. In
  \bibinfo{booktitle}{\emph{Advances in neural information processing
  systems}}. \bibinfo{pages}{2672--2680}.
\newblock


\bibitem[\protect\citeauthoryear{Han, Wu, Wu, Yu, and Davis}{Han
  et~al\mbox{.}}{2018}]%
        {han2018viton}
\bibfield{author}{\bibinfo{person}{Xintong Han}, \bibinfo{person}{Zuxuan Wu},
  \bibinfo{person}{Zhe Wu}, \bibinfo{person}{Ruichi Yu}, {and}
  \bibinfo{person}{Larry~S Davis}.} \bibinfo{year}{2018}\natexlab{}.
\newblock \showarticletitle{Viton: An image-based virtual try-on network}. In
  \bibinfo{booktitle}{\emph{Proceedings of the IEEE conference on computer
  vision and pattern recognition}}. \bibinfo{pages}{7543--7552}.
\newblock


\bibitem[\protect\citeauthoryear{He, Zuo, Kan, Shan, and Chen}{He
  et~al\mbox{.}}{2019}]%
        {he2019attgan}
\bibfield{author}{\bibinfo{person}{Zhenliang He}, \bibinfo{person}{Wangmeng
  Zuo}, \bibinfo{person}{Meina Kan}, \bibinfo{person}{Shiguang Shan}, {and}
  \bibinfo{person}{Xilin Chen}.} \bibinfo{year}{2019}\natexlab{}.
\newblock \showarticletitle{Attgan: Facial attribute editing by only changing
  what you want}.
\newblock \bibinfo{journal}{\emph{IEEE Transactions on Image Processing}}
  \bibinfo{volume}{28}, \bibinfo{number}{11} (\bibinfo{year}{2019}),
  \bibinfo{pages}{5464--5478}.
\newblock


\bibitem[\protect\citeauthoryear{Isola, Zhu, Zhou, and Efros}{Isola
  et~al\mbox{.}}{2017}]%
        {isola2017image}
\bibfield{author}{\bibinfo{person}{Phillip Isola}, \bibinfo{person}{Jun-Yan
  Zhu}, \bibinfo{person}{Tinghui Zhou}, {and} \bibinfo{person}{Alexei~A
  Efros}.} \bibinfo{year}{2017}\natexlab{}.
\newblock \showarticletitle{Image-to-image translation with conditional
  adversarial networks}. In \bibinfo{booktitle}{\emph{Proceedings of the IEEE
  conference on computer vision and pattern recognition}}.
  \bibinfo{pages}{1125--1134}.
\newblock


\bibitem[\protect\citeauthoryear{Jiang and Fu}{Jiang and Fu}{2017}]%
        {jiang2017fashion}
\bibfield{author}{\bibinfo{person}{Shuhui Jiang} {and} \bibinfo{person}{Yun
  Fu}.} \bibinfo{year}{2017}\natexlab{}.
\newblock \showarticletitle{Fashion Style Generator.}. In
  \bibinfo{booktitle}{\emph{IJCAI}}. \bibinfo{pages}{3721--3727}.
\newblock


\bibitem[\protect\citeauthoryear{Kalantidis, Kennedy, and Li}{Kalantidis
  et~al\mbox{.}}{2013}]%
        {kalantidis2013getting}
\bibfield{author}{\bibinfo{person}{Yannis Kalantidis}, \bibinfo{person}{Lyndon
  Kennedy}, {and} \bibinfo{person}{Li-Jia Li}.}
  \bibinfo{year}{2013}\natexlab{}.
\newblock \showarticletitle{Getting the look: clothing recognition and
  segmentation for automatic product suggestions in everyday photos}. In
  \bibinfo{booktitle}{\emph{Proceedings of the 3rd ACM conference on
  International conference on multimedia retrieval}}.
  \bibinfo{pages}{105--112}.
\newblock


\bibitem[\protect\citeauthoryear{Kaneko, Hiramatsu, and Kashino}{Kaneko
  et~al\mbox{.}}{2017}]%
        {kaneko2017generative}
\bibfield{author}{\bibinfo{person}{Takuhiro Kaneko}, \bibinfo{person}{Kaoru
  Hiramatsu}, {and} \bibinfo{person}{Kunio Kashino}.}
  \bibinfo{year}{2017}\natexlab{}.
\newblock \showarticletitle{Generative attribute controller with conditional
  filtered generative adversarial networks}. In
  \bibinfo{booktitle}{\emph{Proceedings of the IEEE conference on computer
  vision and pattern recognition}}. \bibinfo{pages}{6089--6098}.
\newblock


\bibitem[\protect\citeauthoryear{Kang, Fang, Wang, and McAuley}{Kang
  et~al\mbox{.}}{2017}]%
        {kang2017visually}
\bibfield{author}{\bibinfo{person}{Wang-Cheng Kang}, \bibinfo{person}{Chen
  Fang}, \bibinfo{person}{Zhaowen Wang}, {and} \bibinfo{person}{Julian
  McAuley}.} \bibinfo{year}{2017}\natexlab{}.
\newblock \showarticletitle{Visually-aware fashion recommendation and design
  with generative image models}. In \bibinfo{booktitle}{\emph{2017 IEEE
  International Conference on Data Mining (ICDM)}}. IEEE,
  \bibinfo{pages}{207--216}.
\newblock


\bibitem[\protect\citeauthoryear{Kovashka, Parikh, and Grauman}{Kovashka
  et~al\mbox{.}}{2012}]%
        {kovashka2012whittlesearch}
\bibfield{author}{\bibinfo{person}{Adriana Kovashka}, \bibinfo{person}{Devi
  Parikh}, {and} \bibinfo{person}{Kristen Grauman}.}
  \bibinfo{year}{2012}\natexlab{}.
\newblock \showarticletitle{Whittlesearch: Image search with relative attribute
  feedback}. In \bibinfo{booktitle}{\emph{2012 IEEE Conference on Computer
  Vision and Pattern Recognition}}. IEEE, \bibinfo{pages}{2973--2980}.
\newblock


\bibitem[\protect\citeauthoryear{Ledig, Theis, Husz{\'a}r, Caballero,
  Cunningham, Acosta, Aitken, Tejani, Totz, Wang, et~al\mbox{.}}{Ledig
  et~al\mbox{.}}{2017}]%
        {ledig2017photo}
\bibfield{author}{\bibinfo{person}{Christian Ledig}, \bibinfo{person}{Lucas
  Theis}, \bibinfo{person}{Ferenc Husz{\'a}r}, \bibinfo{person}{Jose
  Caballero}, \bibinfo{person}{Andrew Cunningham}, \bibinfo{person}{Alejandro
  Acosta}, \bibinfo{person}{Andrew Aitken}, \bibinfo{person}{Alykhan Tejani},
  \bibinfo{person}{Johannes Totz}, \bibinfo{person}{Zehan Wang},
  {et~al\mbox{.}}} \bibinfo{year}{2017}\natexlab{}.
\newblock \showarticletitle{Photo-realistic single image super-resolution using
  a generative adversarial network}. In \bibinfo{booktitle}{\emph{Proceedings
  of the IEEE conference on computer vision and pattern recognition}}.
  \bibinfo{pages}{4681--4690}.
\newblock


\bibitem[\protect\citeauthoryear{Liang, Lin, Yang, Luo, Huang, and Yan}{Liang
  et~al\mbox{.}}{2016}]%
        {liang2016clothes}
\bibfield{author}{\bibinfo{person}{Xiaodan Liang}, \bibinfo{person}{Liang Lin},
  \bibinfo{person}{Wei Yang}, \bibinfo{person}{Ping Luo},
  \bibinfo{person}{Junshi Huang}, {and} \bibinfo{person}{Shuicheng Yan}.}
  \bibinfo{year}{2016}\natexlab{}.
\newblock \showarticletitle{Clothes co-parsing via joint image segmentation and
  labeling with application to clothing retrieval}.
\newblock \bibinfo{journal}{\emph{IEEE Transactions on Multimedia}}
  \bibinfo{volume}{18}, \bibinfo{number}{6} (\bibinfo{year}{2016}),
  \bibinfo{pages}{1175--1186}.
\newblock


\bibitem[\protect\citeauthoryear{Lin, Ren, Chen, Ren, Ma, and de~Rijke}{Lin
  et~al\mbox{.}}{2018}]%
        {lin2018explainable}
\bibfield{author}{\bibinfo{person}{Yujie Lin}, \bibinfo{person}{Pengjie Ren},
  \bibinfo{person}{Zhumin Chen}, \bibinfo{person}{Zhaochun Ren},
  \bibinfo{person}{Jun Ma}, {and} \bibinfo{person}{Maarten de Rijke}.}
  \bibinfo{year}{2018}\natexlab{}.
\newblock \showarticletitle{Explainable fashion recommendation with joint
  outfit matching and comment generation}.
\newblock \bibinfo{journal}{\emph{arXiv preprint arXiv:1806.08977}}
  \bibinfo{volume}{2} (\bibinfo{year}{2018}).
\newblock


\bibitem[\protect\citeauthoryear{Liu, Luo, Qiu, Wang, and Tang}{Liu
  et~al\mbox{.}}{2016}]%
        {liu2016deepfashion}
\bibfield{author}{\bibinfo{person}{Ziwei Liu}, \bibinfo{person}{Ping Luo},
  \bibinfo{person}{Shi Qiu}, \bibinfo{person}{Xiaogang Wang}, {and}
  \bibinfo{person}{Xiaoou Tang}.} \bibinfo{year}{2016}\natexlab{}.
\newblock \showarticletitle{Deepfashion: Powering robust clothes recognition
  and retrieval with rich annotations}. In
  \bibinfo{booktitle}{\emph{Proceedings of the IEEE conference on computer
  vision and pattern recognition}}. \bibinfo{pages}{1096--1104}.
\newblock


\bibitem[\protect\citeauthoryear{Luc, Couprie, Chintala, and Verbeek}{Luc
  et~al\mbox{.}}{2016}]%
        {luc2016semantic}
\bibfield{author}{\bibinfo{person}{Pauline Luc}, \bibinfo{person}{Camille
  Couprie}, \bibinfo{person}{Soumith Chintala}, {and} \bibinfo{person}{Jakob
  Verbeek}.} \bibinfo{year}{2016}\natexlab{}.
\newblock \showarticletitle{Semantic segmentation using adversarial networks}.
\newblock \bibinfo{journal}{\emph{arXiv preprint arXiv:1611.08408}}
  (\bibinfo{year}{2016}).
\newblock


\bibitem[\protect\citeauthoryear{Luo, Cai, Zhang, Xu, et~al\mbox{.}}{Luo
  et~al\mbox{.}}{2018}]%
        {luo2018multivariate}
\bibfield{author}{\bibinfo{person}{Yonghong Luo}, \bibinfo{person}{Xiangrui
  Cai}, \bibinfo{person}{Ying Zhang}, \bibinfo{person}{Jun Xu},
  {et~al\mbox{.}}} \bibinfo{year}{2018}\natexlab{}.
\newblock \showarticletitle{Multivariate time series imputation with generative
  adversarial networks}. In \bibinfo{booktitle}{\emph{Advances in Neural
  Information Processing Systems}}. \bibinfo{pages}{1596--1607}.
\newblock


\bibitem[\protect\citeauthoryear{Ma, Jia, Sun, Schiele, Tuytelaars, and
  Van~Gool}{Ma et~al\mbox{.}}{2017}]%
        {ma2017pose}
\bibfield{author}{\bibinfo{person}{Liqian Ma}, \bibinfo{person}{Xu Jia},
  \bibinfo{person}{Qianru Sun}, \bibinfo{person}{Bernt Schiele},
  \bibinfo{person}{Tinne Tuytelaars}, {and} \bibinfo{person}{Luc Van~Gool}.}
  \bibinfo{year}{2017}\natexlab{}.
\newblock \showarticletitle{Pose guided person image generation}. In
  \bibinfo{booktitle}{\emph{Advances in Neural Information Processing
  Systems}}. \bibinfo{pages}{406--416}.
\newblock


\bibitem[\protect\citeauthoryear{Mirza and Osindero}{Mirza and
  Osindero}{2014}]%
        {mirza2014conditional}
\bibfield{author}{\bibinfo{person}{Mehdi Mirza} {and} \bibinfo{person}{Simon
  Osindero}.} \bibinfo{year}{2014}\natexlab{}.
\newblock \showarticletitle{Conditional generative adversarial nets}.
\newblock \bibinfo{journal}{\emph{arXiv preprint arXiv:1411.1784}}
  (\bibinfo{year}{2014}).
\newblock


\bibitem[\protect\citeauthoryear{Nie, Narodytska, and Patel}{Nie
  et~al\mbox{.}}{2018}]%
        {nie2018relgan}
\bibfield{author}{\bibinfo{person}{Weili Nie}, \bibinfo{person}{Nina
  Narodytska}, {and} \bibinfo{person}{Ankit Patel}.}
  \bibinfo{year}{2018}\natexlab{}.
\newblock \showarticletitle{Relgan: Relational generative adversarial networks
  for text generation}.
\newblock  (\bibinfo{year}{2018}).
\newblock


\bibitem[\protect\citeauthoryear{Odena, Olah, and Shlens}{Odena
  et~al\mbox{.}}{2017}]%
        {odena2017conditional}
\bibfield{author}{\bibinfo{person}{Augustus Odena},
  \bibinfo{person}{Christopher Olah}, {and} \bibinfo{person}{Jonathon Shlens}.}
  \bibinfo{year}{2017}\natexlab{}.
\newblock \showarticletitle{Conditional image synthesis with auxiliary
  classifier gans}. In \bibinfo{booktitle}{\emph{Proceedings of the 34th
  International Conference on Machine Learning-Volume 70}}. JMLR. org,
  \bibinfo{pages}{2642--2651}.
\newblock


\bibitem[\protect\citeauthoryear{Perarnau, Van De~Weijer, Raducanu, and
  {\'A}lvarez}{Perarnau et~al\mbox{.}}{2016}]%
        {perarnau2016invertible}
\bibfield{author}{\bibinfo{person}{Guim Perarnau}, \bibinfo{person}{Joost Van
  De~Weijer}, \bibinfo{person}{Bogdan Raducanu}, {and} \bibinfo{person}{Jose~M
  {\'A}lvarez}.} \bibinfo{year}{2016}\natexlab{}.
\newblock \showarticletitle{Invertible conditional gans for image editing}.
\newblock \bibinfo{journal}{\emph{arXiv preprint arXiv:1611.06355}}
  (\bibinfo{year}{2016}).
\newblock


\bibitem[\protect\citeauthoryear{Ping, Wu, Ding, and Yuan}{Ping
  et~al\mbox{.}}{2019}]%
        {ping2019fashion}
\bibfield{author}{\bibinfo{person}{Qing Ping}, \bibinfo{person}{Bing Wu},
  \bibinfo{person}{Wanying Ding}, {and} \bibinfo{person}{Jiangbo Yuan}.}
  \bibinfo{year}{2019}\natexlab{}.
\newblock \showarticletitle{Fashion-AttGAN: Attribute-aware fashion editing
  with multi-objective GAN}. In \bibinfo{booktitle}{\emph{Proceedings of the
  IEEE Conference on Computer Vision and Pattern Recognition Workshops}}.
  \bibinfo{pages}{0--0}.
\newblock


\bibitem[\protect\citeauthoryear{Raj, Sangkloy, Chang, Lu, Ceylan, and
  Hays}{Raj et~al\mbox{.}}{2018}]%
        {raj2018swapnet}
\bibfield{author}{\bibinfo{person}{Amit Raj}, \bibinfo{person}{Patsorn
  Sangkloy}, \bibinfo{person}{Huiwen Chang}, \bibinfo{person}{Jingwan Lu},
  \bibinfo{person}{Duygu Ceylan}, {and} \bibinfo{person}{James Hays}.}
  \bibinfo{year}{2018}\natexlab{}.
\newblock \showarticletitle{Swapnet: Garment transfer in single view images}.
  In \bibinfo{booktitle}{\emph{Proceedings of the European Conference on
  Computer Vision (ECCV)}}. \bibinfo{pages}{666--682}.
\newblock


\bibitem[\protect\citeauthoryear{Reed, Akata, Yan, Logeswaran, Schiele, and
  Lee}{Reed et~al\mbox{.}}{2016}]%
        {reed2016generative}
\bibfield{author}{\bibinfo{person}{Scott Reed}, \bibinfo{person}{Zeynep Akata},
  \bibinfo{person}{Xinchen Yan}, \bibinfo{person}{Lajanugen Logeswaran},
  \bibinfo{person}{Bernt Schiele}, {and} \bibinfo{person}{Honglak Lee}.}
  \bibinfo{year}{2016}\natexlab{}.
\newblock \showarticletitle{Generative adversarial text to image synthesis}.
\newblock \bibinfo{journal}{\emph{arXiv preprint arXiv:1605.05396}}
  (\bibinfo{year}{2016}).
\newblock


\bibitem[\protect\citeauthoryear{Schroff, Kalenichenko, and Philbin}{Schroff
  et~al\mbox{.}}{2015}]%
        {schroff2015facenet}
\bibfield{author}{\bibinfo{person}{Florian Schroff}, \bibinfo{person}{Dmitry
  Kalenichenko}, {and} \bibinfo{person}{James Philbin}.}
  \bibinfo{year}{2015}\natexlab{}.
\newblock \showarticletitle{Facenet: A unified embedding for face recognition
  and clustering}. In \bibinfo{booktitle}{\emph{Proceedings of the IEEE
  conference on computer vision and pattern recognition}}.
  \bibinfo{pages}{815--823}.
\newblock


\bibitem[\protect\citeauthoryear{Shen and Liu}{Shen and Liu}{2017}]%
        {shen2017learning}
\bibfield{author}{\bibinfo{person}{Wei Shen} {and} \bibinfo{person}{Rujie
  Liu}.} \bibinfo{year}{2017}\natexlab{}.
\newblock \showarticletitle{Learning residual images for face attribute
  manipulation}. In \bibinfo{booktitle}{\emph{Proceedings of the IEEE
  conference on computer vision and pattern recognition}}.
  \bibinfo{pages}{4030--4038}.
\newblock


\bibitem[\protect\citeauthoryear{Simo-Serra, Fidler, Moreno-Noguer, and
  Urtasun}{Simo-Serra et~al\mbox{.}}{2015}]%
        {simo2015neuroaesthetics}
\bibfield{author}{\bibinfo{person}{Edgar Simo-Serra}, \bibinfo{person}{Sanja
  Fidler}, \bibinfo{person}{Francesc Moreno-Noguer}, {and}
  \bibinfo{person}{Raquel Urtasun}.} \bibinfo{year}{2015}\natexlab{}.
\newblock \showarticletitle{Neuroaesthetics in fashion: Modeling the perception
  of fashionability}. In \bibinfo{booktitle}{\emph{Proceedings of the IEEE
  Conference on Computer Vision and Pattern Recognition}}.
  \bibinfo{pages}{869--877}.
\newblock


\bibitem[\protect\citeauthoryear{Xian, Sangkloy, Agrawal, Raj, Lu, Fang, Yu,
  and Hays}{Xian et~al\mbox{.}}{2018}]%
        {xian2018texturegan}
\bibfield{author}{\bibinfo{person}{Wenqi Xian}, \bibinfo{person}{Patsorn
  Sangkloy}, \bibinfo{person}{Varun Agrawal}, \bibinfo{person}{Amit Raj},
  \bibinfo{person}{Jingwan Lu}, \bibinfo{person}{Chen Fang},
  \bibinfo{person}{Fisher Yu}, {and} \bibinfo{person}{James Hays}.}
  \bibinfo{year}{2018}\natexlab{}.
\newblock \showarticletitle{Texturegan: Controlling deep image synthesis with
  texture patches}. In \bibinfo{booktitle}{\emph{Proceedings of the IEEE
  Conference on Computer Vision and Pattern Recognition}}.
  \bibinfo{pages}{8456--8465}.
\newblock


\bibitem[\protect\citeauthoryear{Xu, Chen, Liu, Chen, Weng, Hong, and Lin}{Xu
  et~al\mbox{.}}{2019a}]%
        {xu2019topology}
\bibfield{author}{\bibinfo{person}{Kaidi Xu}, \bibinfo{person}{Hongge Chen},
  \bibinfo{person}{Sijia Liu}, \bibinfo{person}{Pin-Yu Chen},
  \bibinfo{person}{Tsui-Wei Weng}, \bibinfo{person}{Mingyi Hong}, {and}
  \bibinfo{person}{Xue Lin}.} \bibinfo{year}{2019}\natexlab{a}.
\newblock \showarticletitle{Topology Attack and Defense for Graph Neural
  Networks: An Optimization Perspective}. In
  \bibinfo{booktitle}{\emph{International Joint Conference on Artificial
  Intelligence (IJCAI)}}.
\newblock


\bibitem[\protect\citeauthoryear{Xu, Liu, Zhang, Sun, Zhao, Fan, Gan, and
  Lin}{Xu et~al\mbox{.}}{2019b}]%
        {xu2019interpreting}
\bibfield{author}{\bibinfo{person}{Kaidi Xu}, \bibinfo{person}{Sijia Liu},
  \bibinfo{person}{Gaoyuan Zhang}, \bibinfo{person}{Mengshu Sun},
  \bibinfo{person}{Pu Zhao}, \bibinfo{person}{Quanfu Fan},
  \bibinfo{person}{Chuang Gan}, {and} \bibinfo{person}{Xue Lin}.}
  \bibinfo{year}{2019}\natexlab{b}.
\newblock \showarticletitle{Interpreting adversarial examples by activation
  promotion and suppression}.
\newblock \bibinfo{journal}{\emph{arXiv preprint arXiv:1904.02057}}
  (\bibinfo{year}{2019}).
\newblock


\bibitem[\protect\citeauthoryear{Xu, Liu, Zhao, Chen, Zhang, Fan, Erdogmus,
  Wang, and Lin}{Xu et~al\mbox{.}}{2019c}]%
        {xu2018structured}
\bibfield{author}{\bibinfo{person}{Kaidi Xu}, \bibinfo{person}{Sijia Liu},
  \bibinfo{person}{Pu Zhao}, \bibinfo{person}{Pin-Yu Chen},
  \bibinfo{person}{Huan Zhang}, \bibinfo{person}{Quanfu Fan},
  \bibinfo{person}{Deniz Erdogmus}, \bibinfo{person}{Yanzhi Wang}, {and}
  \bibinfo{person}{Xue Lin}.} \bibinfo{year}{2019}\natexlab{c}.
\newblock \showarticletitle{Structured Adversarial Attack: Towards General
  Implementation and Better Interpretability}. In
  \bibinfo{booktitle}{\emph{International Conference on Learning
  Representations}}.
\newblock


\bibitem[\protect\citeauthoryear{Xu, Zhang, Liu, Fan, Sun, Chen, Chen, Wang,
  and Lin}{Xu et~al\mbox{.}}{2020}]%
        {xu2019evading}
\bibfield{author}{\bibinfo{person}{Kaidi Xu}, \bibinfo{person}{Gaoyuan Zhang},
  \bibinfo{person}{Sijia Liu}, \bibinfo{person}{Quanfu Fan},
  \bibinfo{person}{Mengshu Sun}, \bibinfo{person}{Hongge Chen},
  \bibinfo{person}{Pin-Yu Chen}, \bibinfo{person}{Yanzhi Wang}, {and}
  \bibinfo{person}{Xue Lin}.} \bibinfo{year}{2020}\natexlab{}.
\newblock \showarticletitle{Adversarial t-shirt! evading person detectors in a
  physical world}. In \bibinfo{booktitle}{\emph{Proceedings of the European
  Conference on Computer Vision (ECCV)}}.
\newblock


\bibitem[\protect\citeauthoryear{Xu, Zhang, Huang, Zhang, Gan, Huang, and
  He}{Xu et~al\mbox{.}}{2018}]%
        {xu2018attngan}
\bibfield{author}{\bibinfo{person}{Tao Xu}, \bibinfo{person}{Pengchuan Zhang},
  \bibinfo{person}{Qiuyuan Huang}, \bibinfo{person}{Han Zhang},
  \bibinfo{person}{Zhe Gan}, \bibinfo{person}{Xiaolei Huang}, {and}
  \bibinfo{person}{Xiaodong He}.} \bibinfo{year}{2018}\natexlab{}.
\newblock \showarticletitle{Attngan: Fine-grained text to image generation with
  attentional generative adversarial networks}. In
  \bibinfo{booktitle}{\emph{Proceedings of the IEEE conference on computer
  vision and pattern recognition}}. \bibinfo{pages}{1316--1324}.
\newblock


\bibitem[\protect\citeauthoryear{Zhang, Kan, Shan, and Chen}{Zhang
  et~al\mbox{.}}{2018}]%
        {zhang2018generative}
\bibfield{author}{\bibinfo{person}{Gang Zhang}, \bibinfo{person}{Meina Kan},
  \bibinfo{person}{Shiguang Shan}, {and} \bibinfo{person}{Xilin Chen}.}
  \bibinfo{year}{2018}\natexlab{}.
\newblock \showarticletitle{Generative adversarial network with spatial
  attention for face attribute editing}. In
  \bibinfo{booktitle}{\emph{Proceedings of the European Conference on Computer
  Vision (ECCV)}}. \bibinfo{pages}{417--432}.
\newblock


\bibitem[\protect\citeauthoryear{Zhang, Xu, Li, Zhang, Wang, Huang, and
  Metaxas}{Zhang et~al\mbox{.}}{2017}]%
        {zhang2017stackgan}
\bibfield{author}{\bibinfo{person}{Han Zhang}, \bibinfo{person}{Tao Xu},
  \bibinfo{person}{Hongsheng Li}, \bibinfo{person}{Shaoting Zhang},
  \bibinfo{person}{Xiaogang Wang}, \bibinfo{person}{Xiaolei Huang}, {and}
  \bibinfo{person}{Dimitris~N Metaxas}.} \bibinfo{year}{2017}\natexlab{}.
\newblock \showarticletitle{Stackgan: Text to photo-realistic image synthesis
  with stacked generative adversarial networks}. In
  \bibinfo{booktitle}{\emph{Proceedings of the IEEE international conference on
  computer vision}}. \bibinfo{pages}{5907--5915}.
\newblock


\bibitem[\protect\citeauthoryear{Zhao, Feng, Wu, and Yan}{Zhao
  et~al\mbox{.}}{2017}]%
        {zhao2017memory}
\bibfield{author}{\bibinfo{person}{Bo Zhao}, \bibinfo{person}{Jiashi Feng},
  \bibinfo{person}{Xiao Wu}, {and} \bibinfo{person}{Shuicheng Yan}.}
  \bibinfo{year}{2017}\natexlab{}.
\newblock \showarticletitle{Memory-augmented attribute manipulation networks
  for interactive fashion search}. In \bibinfo{booktitle}{\emph{Proceedings of
  the IEEE Conference on Computer Vision and Pattern Recognition}}.
  \bibinfo{pages}{1520--1528}.
\newblock


\bibitem[\protect\citeauthoryear{Zhu, Zhang, Pathak, Darrell, Efros, Wang, and
  Shechtman}{Zhu et~al\mbox{.}}{2017b}]%
        {zhu2017toward}
\bibfield{author}{\bibinfo{person}{Jun-Yan Zhu}, \bibinfo{person}{Richard
  Zhang}, \bibinfo{person}{Deepak Pathak}, \bibinfo{person}{Trevor Darrell},
  \bibinfo{person}{Alexei~A Efros}, \bibinfo{person}{Oliver Wang}, {and}
  \bibinfo{person}{Eli Shechtman}.} \bibinfo{year}{2017}\natexlab{b}.
\newblock \showarticletitle{Toward multimodal image-to-image translation}. In
  \bibinfo{booktitle}{\emph{Advances in neural information processing
  systems}}. \bibinfo{pages}{465--476}.
\newblock


\bibitem[\protect\citeauthoryear{Zhu, Urtasun, Fidler, Lin, and Change~Loy}{Zhu
  et~al\mbox{.}}{2017a}]%
        {zhu2017your}
\bibfield{author}{\bibinfo{person}{Shizhan Zhu}, \bibinfo{person}{Raquel
  Urtasun}, \bibinfo{person}{Sanja Fidler}, \bibinfo{person}{Dahua Lin}, {and}
  \bibinfo{person}{Chen Change~Loy}.} \bibinfo{year}{2017}\natexlab{a}.
\newblock \showarticletitle{Be your own prada: Fashion synthesis with
  structural coherence}. In \bibinfo{booktitle}{\emph{Proceedings of the IEEE
  international conference on computer vision}}. \bibinfo{pages}{1680--1688}.
\newblock


\end{thebibliography}


\end{document}